\PassOptionsToPackage{unicode}{hyperref}
\PassOptionsToPackage{hyphens}{url}
\documentclass[
]{article}
\title{Deep neural networks as nested dynamical systems}
\author{David I. Spivak \and Timothy Hosgood}

\usepackage{amsmath,amssymb}
\usepackage{lmodern}
\usepackage{iftex}
\ifPDFTeX
  \usepackage[T1]{fontenc}
  \usepackage[utf8]{inputenc}
  \usepackage{textcomp} 
\else 
  \usepackage{unicode-math}
  \defaultfontfeatures{Scale=MatchLowercase}
  \defaultfontfeatures[\rmfamily]{Ligatures=TeX,Scale=1}
\fi
\IfFileExists{upquote.sty}{\usepackage{upquote}}{}
\IfFileExists{microtype.sty}{
  \usepackage[]{microtype}
  \UseMicrotypeSet[protrusion]{basicmath} 
}{}
\makeatletter
\@ifundefined{KOMAClassName}{
  \IfFileExists{parskip.sty}{%
    \usepackage{parskip}
  }{
    \setlength{\parindent}{0pt}
    \setlength{\parskip}{6pt plus 2pt minus 1pt}}
}{
  \KOMAoptions{parskip=half}}
\makeatother
\usepackage{xcolor}
\IfFileExists{xurl.sty}{\usepackage{xurl}}{} 
\IfFileExists{bookmark.sty}{\usepackage{bookmark}}{\usepackage{hyperref}}
\hypersetup{
  pdftitle={Deep neural networks as nested dynamical systems},
  pdfauthor={David I. Spivak; Timothy Hosgood},
  hidelinks,
  pdfcreator={LaTeX via pandoc}}
\usepackage{longtable,booktabs,array}
\usepackage{calc} 
\usepackage{etoolbox}
\makeatletter
\patchcmd\longtable{\par}{\if@noskipsec\mbox{}\fi\par}{}{}
\makeatother
\IfFileExists{footnotehyper.sty}{\usepackage{footnotehyper}}{\usepackage{footnote}}
\makesavenoteenv{longtable}
\usepackage{graphicx}
\makeatletter
\def\maxwidth{\ifdim\Gin@nat@width>\linewidth\linewidth\else\Gin@nat@width\fi}
\def\maxheight{\ifdim\Gin@nat@height>\textheight\textheight\else\Gin@nat@height\fi}
\makeatother
\setkeys{Gin}{width=\maxwidth,height=\maxheight,keepaspectratio}
\makeatletter
\def\fps@figure{htbp}
\makeatother
\providecommand{\tightlist}{%
  \setlength{\itemsep}{0pt}\setlength{\parskip}{0pt}}
\usepackage[margin=1.2in]{geometry}
\usepackage[cal=esstix]{mathalfa}
\usepackage{amssymb}
\usepackage{pifont}

\usepackage{float}
\makeatletter
\def\fps@figure{H}
\makeatother
\ifLuaTeX
  \usepackage{selnolig}  
\fi

\begin{document}
\maketitle

An oft-repeated aphorism is that ``category theory is mathematics for
making analogies precise'', and there is truth in this. Indeed, one
of the reasons for category theory's ubiquity in modern mathematics is
its ability to turn vague-sounding stories into formal concrete
mathematics, and it is a specific example of this phenomenon that we are
going to discuss here.

There is an analogy that is often made between deep neural networks and
actual brains, suggested by the nomenclature itself: the ``neurons'' in
deep neural networks should correspond to neurons (or nerve cells, to
avoid confusion) in the brain. We claim, however, that this analogy
doesn't even type check: \emph{it is structurally flawed}. In agreement
with the slightly glib summary of Hebbian learning as ``cells that fire
together wire together'', this article makes the case that the analogy
should be different. Since the ``neurons'' in deep neural networks are
managing the changing weights, they are more akin to the
\emph{synapses} in the brain; instead, it is the \emph{wires} in deep neural
networks that are more like nerve cells, in that they are what cause the
information to flow. An intuition that nerve cells seem like
more than mere wires is exactly right, and is justified by a precise category-theoretic analogy which we will explore
in this article. Throughout, we will continue to highlight the error in equating artificial neurons with nerve cells by
leaving ``neuron'' in quotes or by calling them artificial neurons.

We will first explain how to view deep neural networks as nested
dynamical systems with a very restricted sort of interaction pattern, and then
explain a more general sort of interaction for dynamical systems that
is useful throughout engineering, but which fails to adapt to changing
circumstances. As mentioned, an analogy is then forced upon us by the
mathematical formalism in which they are both embedded. We call the
resulting encompassing generalization \textbf{deeply interacting
learning systems}: they have complex interaction as in control theory, but
adaptation to circumstances as in deep neural networks.

\hypertarget{section}{%
\subsection{1}\label{section}}

\begin{figure}
\centering
\includegraphics[width=1\textwidth,height=\textheight]{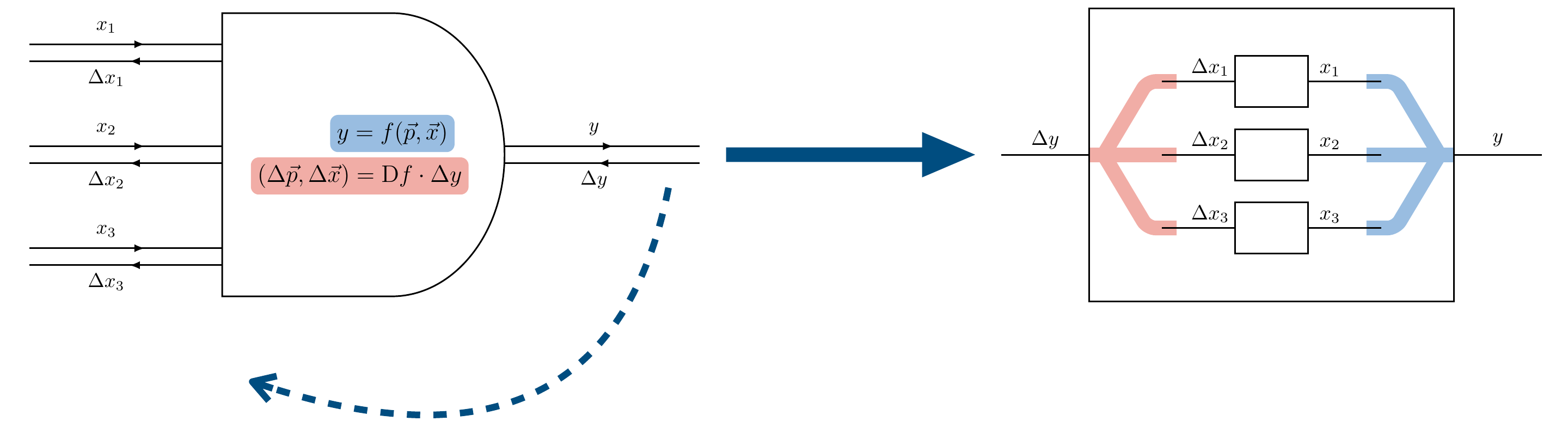}
\caption{\emph{By grabbing hold of the wire labelled \(\Delta y\) and
dragging it along the dotted arrow, we can ``unfold'' our artificial neuron (shown left) into an
interaction diagram (shown right), so that the inputs and output to the artificial neuron turn into interior boxes and exterior box respectively.}\label{figure:unfolding}}
\end{figure}

The process of training a \textbf{deep neural network} (DNN) can be
described as follows: an input-output pair is given (this is a training
datum); the DNN uses its current parameters to push the given input from
the start to the end of the network; the DNN compares the final result
of this forward pass against the given output, and propagates the error
backward through the network, updating its parameters as it goes; the whole
process is repeated many times. DNNs are commonly drawn as networks of
artificial neurons; it is through the wires connecting the ``neurons'' that
information is passed, and information is passed in both directions.
Because of this, our first slight modification to these pictures is to
double each of the wires, resulting in two unidirectional paths instead
of one bidirectional path, as shown in the left-hand side of Figure
\ref{figure:unfolding}. Now the wires running from left to right
(labelled \(x_i\), \(y_i\), etc.) correspond to the forward pass, and
the wires running from right to left (labelled \(\Delta x_i\),
\(\Delta y_i\), etc.) correspond to the backward pass.

The most important mental move required to understand the mathematical
analogy we are building is as follows. Imagine taking a single
``neuron'', along with its input and output wires, and ``unfolding'' it
into a different shaped diagram, as shown in Figure
\ref{figure:unfolding}. Doing so results in a trivial example of
something called a \textbf{interaction diagram}, and, as we will
discuss, these objects give a very useful way of describing
\textbf{interacting dynamical systems} (IDSs). 

\begin{figure}
\centering
\includegraphics[width=0.9\textwidth,height=\textheight]{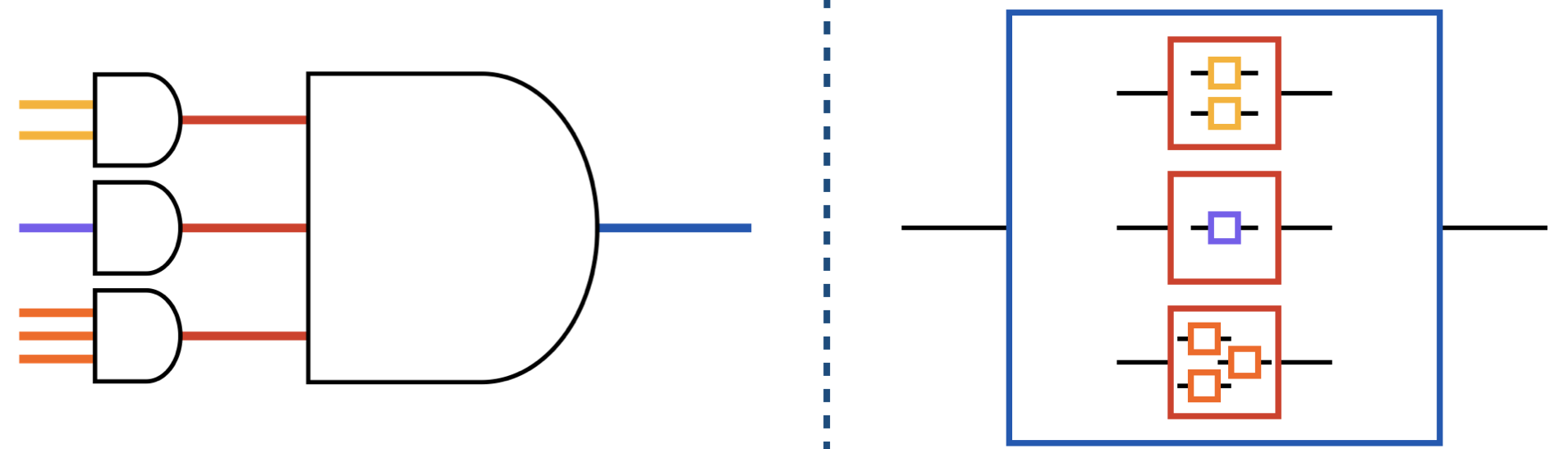}
\caption{\emph{In the case of a multiple artificial neurons, the unfolding
procedure turns left-to-right hierarchy into inside-to-outside
hierarchy. We indicate the idea here via counts and colors.}\label{figure:unfolding-multiple}}
\end{figure}

Generalising the above
procedure to multiple ``neurons'', as in Figure
\ref{figure:unfolding-multiple}, we see how left-to-right composition
turns into so-called \textbf{operadic} composition, given by nesting
interaction diagrams inside other interaction diagrams.

Before going any further though, let us take a moment to pause and gain
some intuition on how to read interaction diagrams. We can think of each
box as a person holding an office---it is important that we really
mean \emph{both} a person \emph{and} the office that they hold; the
person can make decisions, and the office has attendant capacities that
allow the person to \emph{abstract} the information they receive from
the world (not just from subordinates, but from their superiors as
well). That is, the person is given \emph{lower-level data} and turns it
into \emph{higher-level data}. We refer to the person as the
\textbf{abstractor} and the function of their office (i.e.~how it
interacts with the rest of the world) as the \textbf{abstraction}.

Now we can understand the interaction diagram for DNNs in Figure
\ref{figure:unfolding} as follows. There are three smaller abstractors
(which we will call ``workers''), whose offices are in the purview of a
higher-level abstractor (which we will refer to as M, for ``manager''),
drawn as an encompassing box. So M receives information from the outside
(the ``input'', which we can think of as \emph{feedback}, as in ``good
job'' or ``make this correction'') and is tasked with calling out some
sort of processed information (the ``output'') back to the outside
world. The way that M does this is by updating its methodology
(namely, how much it listens to, or trusts, each of its subordinate workers;
the associated \textbf{weights and biases}), as well as passing along to each
worker W the part of the input that corresponds to W's office
(that is, the feedback specific to that worker). This recurses inwards:
each worker does whatever it is that they have been trained to do, sends
feedback to and then listens to subordinates, and shouts out the results to its superiors. In proportion to how loudly each worker shouts, \emph{and}
according to M's current weights and biases, M receives the three
outputs and combines them to give the final output, which it shouts out.
Note that M is also shouting things out here---this whole system can
recurse outwards as well, with M being subordinate to some other,
higher-level managers. In DNNs, the levels of this hierarchy are called \emph{layers}.

Of course, this scenario where workers are competing in a shouting match
to get their manager to listen to them is not really the description of
an ideal working environment. Most notably, \emph{the workers in DNNs
never talk to each other or collaborate}. And yet this is how almost all
neural networks work today; the only thing an artificial neuron can do
is add up incoming signals according to weights and biases. Note that even
weight tying, as in convolutional or recurrent neural networks, does not
address the point we're making here. Certain loss functions---as seen in
physics-inspired neural networks, for example---do make a small amount
of headway, but the breadth of what is possible here does not seem to be
known in the machine learning community. To remedy this, we will
explain in the next section an idea that \emph{is} well-known in the
control theory community.

\begin{figure}
\centering
\includegraphics[width=\textwidth]{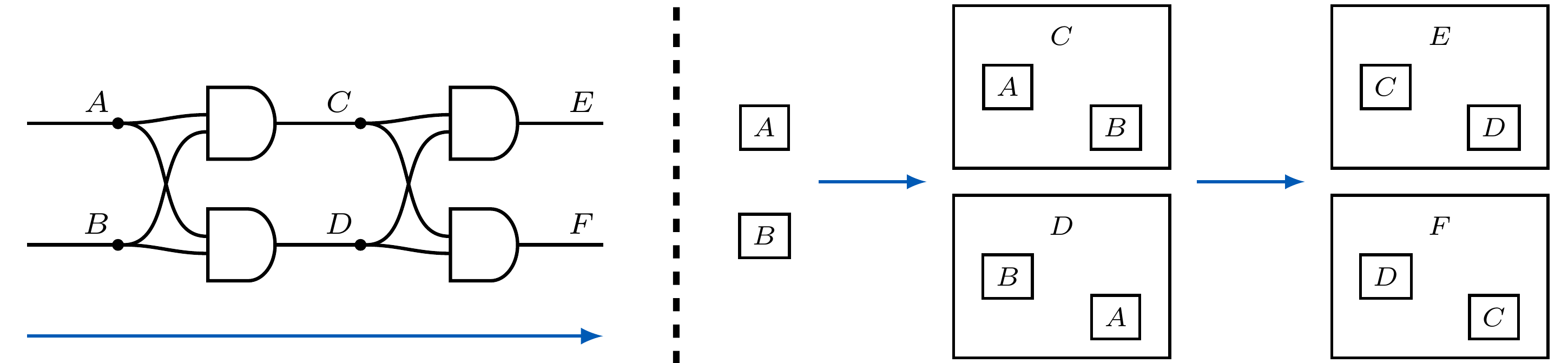}
\caption{A usual deep neural network (shown left) corresponds to an interaction diagram with trivial internal wiring (shown right).\label{figure:boring-wiring-diagram}}
\end{figure}

\begin{figure}
\centering
\includegraphics[width=0.5\textwidth]{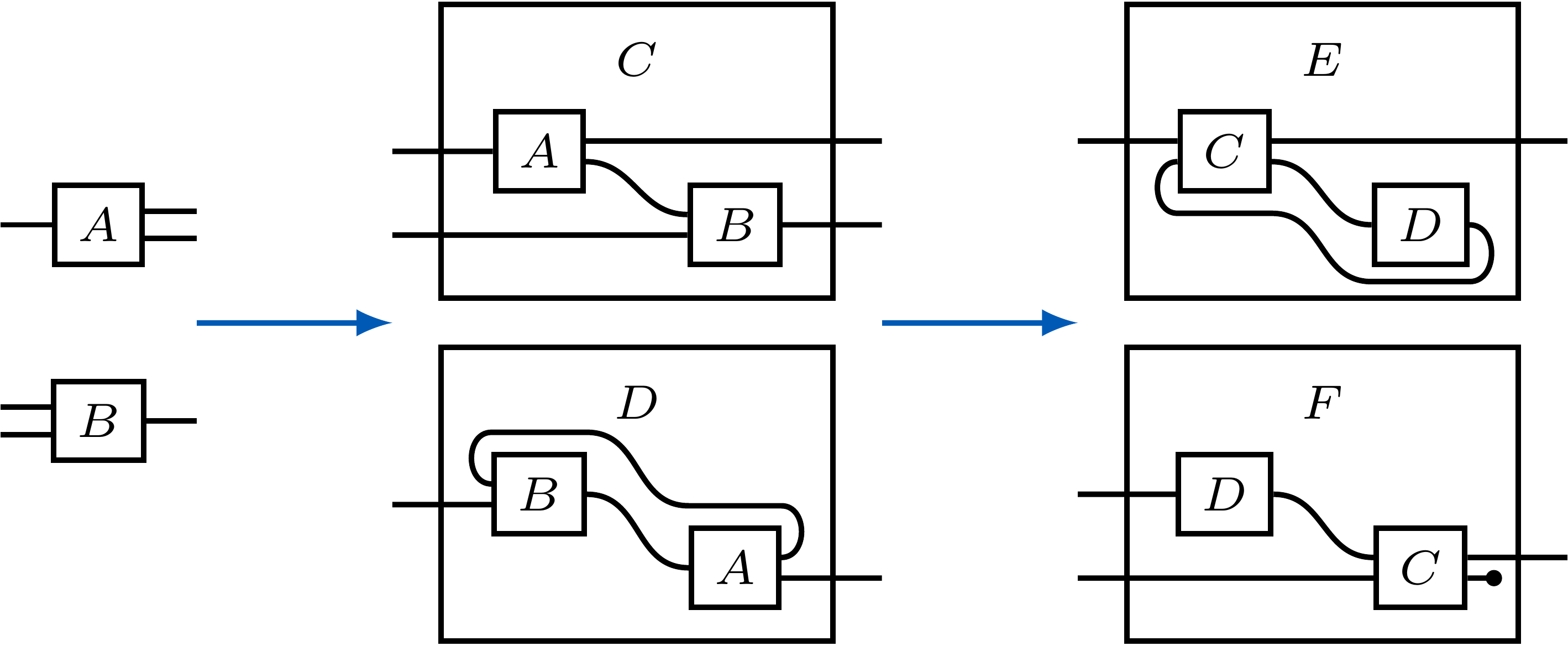}
\caption{An alternative version of the interaction diagram in Figure~\ref{figure:boring-wiring-diagram}, with non-trivial internal wiring.\label{figure:exciting-wiring-diagram}}
\end{figure}

Just to be clear though, this problem is not simply due to the fact that
we are considering one single artificial neuron, or that there is no
feedback in the typical diagram, as one might imagine is remedied by recurrent
neural nets. Again, having loops in a typical DNN diagram does not type check with the remedy we're proposing, because nerve cells are playing the role of wires
in these typical diagrams. The unfolded diagrams give a pictorial representation that clarifies how to
see the nerve cells (which were the wires and are now the boxes) as interacting. Even in the case of multiple ``neurons"
(cf.~Figure~\ref{figure:boring-wiring-diagram}), although there are more
levels of nesting, there is still no ``peer-to-peer communication''. Thus we are ready to pose the question, \emph{``what if we allowed our
DNNs to have internal wiring?''} (cf.~Figure
\ref{figure:exciting-wiring-diagram}).

\hypertarget{section-1}{%
\subsection{2}\label{section-1}}

Let us now take some time to discuss interaction diagrams---a
generalization of something called wiring diagrams or block diagrams,
commonly found in control theory---and how they describe interacting
dynamical systems. Here we will see that, although such diagrams gain a great deal of generality by having non-trivial wirings, they have \emph{no notion} of adaptivity, as found in DNNs.

Consider a computer processor. It consists of many (many) small parts,
and these small parts are made up of even smaller parts: the circuit
board has adders, which are made up of logic gates, which are made from
transistors, and so on. Using the language from before, we can say that
the adder is an {abstractor} of the logic gates, and the logic
gates themselves are {abstractors} of the transistors. Some
abstractions are more useful than others, so people have spent a great
deal of effort in finding the right abstractions, such as \texttt{NOT},
\texttt{AND}, \texttt{OR}, etc. At each level of this hierarchy,
\emph{the parts interact with each other}, resulting in the processes
which combine to give the functionality of the processor as a whole. But
then the processor itself forms a smaller part of a larger system, such
as a computer, or a mobile phone, and these in turn might combine to
form networks with their own emergent behaviour. This is an example of
\textbf{operadic composition} of systems, which is the formal way of specifying how simpler abstractions can be combined to form higher level ones.

\begin{figure}
\centering
\includegraphics[width=0.3\textwidth,height=\textheight]{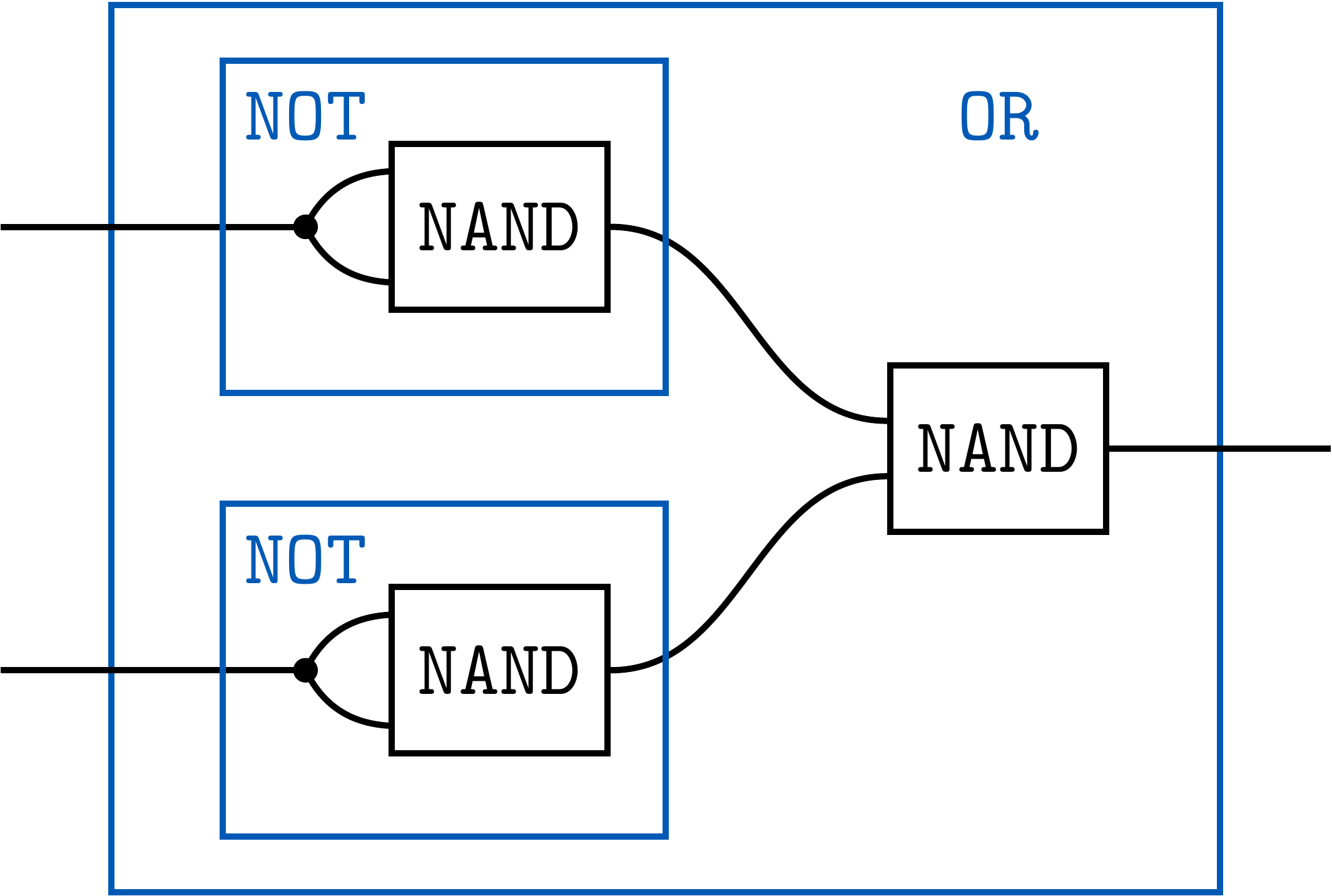}
\caption{\emph{Constructing the logical \texttt{OR} gate from
\texttt{NAND} gates. Here, the wires describe how the inputs and outputs
of the system are interconnected, and the blue boxes demarcate
(sub)systems, with the \texttt{NOT} gate being an \textbf{intermediate
abstraction} between the lower-level \texttt{NAND}s and the higher-level
\texttt{OR}.}\label{figure:operadic-composition}}
\end{figure}

\begin{figure}
\centering
\includegraphics[width=1\textwidth,height=\textheight]{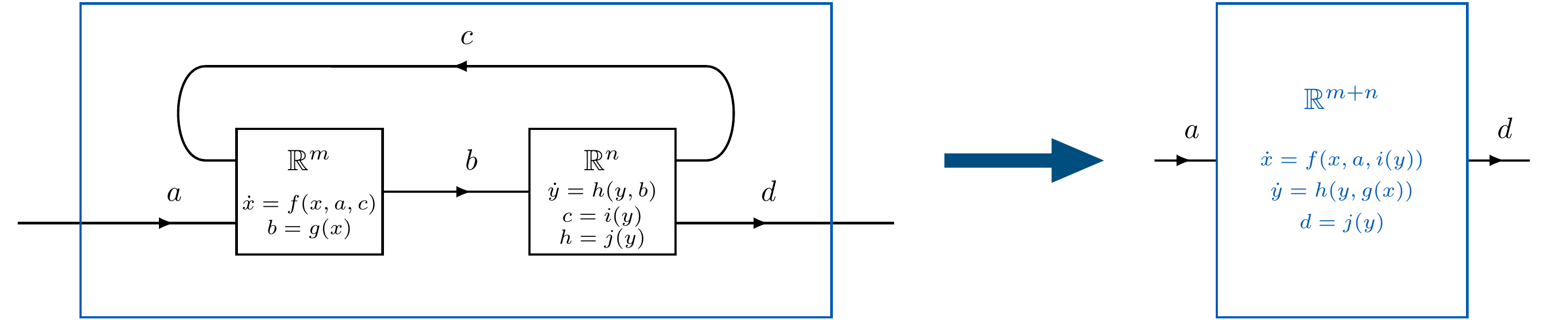}
\caption{\emph{Variable sharing allows us to compose open dynamical
systems.}\label{figure:ODE-composition}}
\end{figure}

We can draw these systems and subsystems using interaction diagrams. For
example, Figure~\ref{figure:operadic-composition} shows how we can
construct the logical \texttt{OR} gate using \texttt{NAND} gates. But
these diagrams are not simply convenient pictorial representation of the
dynamical systems which they describe; instead, using category theory,
we can regard these diagrams as \emph{formal mathematical objects},
whose combinatorics is the subject of mathematical statements and
proofs. Figure~\ref{figure:ODE-composition} gives an example of this,
showing that we can embed a certain sort of mathematical objects (here, systems of ODEs) into
these diagrams, and get another such object (a combined ODE) as a result.

In this diagrammatic approach to studying interacting
dynamical systems, the base-level systems are dynamic and
changing, but the formalism seems to insist that \emph{their interaction
pattern is itself fixed in time}. Indeed, if we imagine the actual
electronic circuit described by Figure
\ref{figure:operadic-composition}, for example, then the connections
would be permanently soldered, and the way in which the subsystems
interact would never change. Of course, from the birth of modern
electronics up until now, fixed interaction patterns have proved very
useful---we're not denying that! But in real life, the wirings between
systems can often change: FPGAs provide one example, but so do more
general scenarios, such as the way that mobile phones move in and out of
range of network coverage, causing reconnection to different cell
towers. We can even consider the problem of studying human society as a
dynamical system, and here the connections are constantly changing: two
people who might be ``wired together'' (e.g.~talking via video call) at
one point in time will then be ``un-wired'' (e.g.~ending the call) at a
later point in time. In the language of interaction diagrams, allowing
such continuous changes in interaction patterns is called
\textbf{dynamic rewiring}, and it's becoming increasingly important to have a mathematical account of.

One of our earlier claims was that DNNs already have the capacity for
such dynamic rewiring. Indeed, this is inherent in the way that these
systems learn: the values of weights and biases change after every batch
of training data. So by unfolding the usual neuron picture into the
interaction diagram picture (see Figure~\ref{figure:unfolding}), and thinking of the nerve cells as the
interior boxes (note that these two changes are enough to repair the
usual analogy), we have allowed our DNNs to have peer-to-peer messaging,
but have perhaps lost the possibility for changing interaction patterns. 

The question then, is the natural one: \emph{``can we have the best of
both worlds, allowing our systems to have (a) non-trivial peer-to-peer
wiring, that (b) can change and adapt through time?''}

\hypertarget{section-2}{%
\subsection{3}\label{section-2}}

Using category theory (or, more specifically, the formalism of
\textbf{polynomial functors}), one can give a positive answer to the
above question. The two generalisations that we have described actually
give one single structure: an \emph{interacting dynamical system with dynamic
rewiring} and a \emph{deep neural network with peer-to-peer messaging} are
simply two different descriptions of the same mathematical object, which we call
\textbf{deeply interactive learning systems} (DILSs). Passing between
these two viewpoints allows us to better understand the analogy between
them.

With DILSs, there is no longer the usual discrete partition of the
learner into ``learning phase'', ``testing phase'', ``implementation
phase''---instead, the system is continuously online, embedded in an
actual world, as is the case in control theory. The usual notion of trainer is replaced simply by prediction error: how well the high-level abstractions actually offer affordances to the system. From the DNN point of
view, ``the current collection of weights and biases'' is generalised to ``the
current interaction pattern between components''. Furthermore, these
interaction patterns are much more collaborative and cooperative than
the simple ``shouting match'' described by weights and biases. It also
helps us to understand the relation between data as it flows through the
DNN, as well as abstraction itself: the lefthand layer corresponds to low-level data (think
``pixels''); the right-hand layer corresponds to high-level data (think ``cat'');
processing data corresponds to creating higher-level abstractions (think
``pixels become curves and features, which become... ears and whiskers, which become a cat''). This corresponds to the movement from interior to exterior boxes of Figure~\ref{figure:exciting-wiring-diagram}, whereas the movement from left to right wires correspond more to going from sensory input to motor output.

\begin{longtable}[]{@{}
  >{\raggedleft\arraybackslash}p{(\columnwidth - 6\tabcolsep) * \real{0.18}}
  >{\centering\arraybackslash}p{(\columnwidth - 6\tabcolsep) * \real{0.27}}
  >{\centering\arraybackslash}p{(\columnwidth - 6\tabcolsep) * \real{0.27}}
  >{\centering\arraybackslash}p{(\columnwidth - 6\tabcolsep) * \real{0.27}}@{}}
\toprule
\begin{minipage}[b]{\linewidth}\raggedleft
\end{minipage} & \begin{minipage}[b]{\linewidth}\centering
Hierarchy direction
\end{minipage} & \begin{minipage}[b]{\linewidth}\centering
Peer-to-peer messaging?
\end{minipage} & \begin{minipage}[b]{\linewidth}\centering
Changeable interaction pattern?
\end{minipage} \\
\midrule
\endhead
DNN & left \(\to\) right & \ding{55} & \ding{51} \\
IDS & inside \(\to\) outside & \ding{51} & \ding{55} \\
DILS & either/both of the above & \ding{51} & \ding{51} \\
\bottomrule
\end{longtable}

Interacting dynamical systems with fixed wiring have been extraordinary
useful over the past-half century, but they are inherently static. The
power of deep neural networks explicitly relies on dynamic
rewiring, but they neglect the possibility of peer-to-peer messaging.
Category theory allows us to combine the strengths of these two
architectures into one more general framework (whose applications are as
yet unexplored), whilst also correcting the structural flaw in usual
analogy between deep neural networks and brain anatomy.

\hypertarget{reading-list}{%
\subsection{Reading list}\label{reading-list}}

\begin{enumerate}
\def\labelenumi{\arabic{enumi}.}
\tightlist
\item
  GSH Cruttwell, B Gavranović, N Ghani, P Wilson, F Zanasi.
  ``Categorical Foundations of Gradient-Based Learning.'' (2021)
  \href{https://arxiv.org/abs/2103.01931}{arXiv: \texttt{2103.01931}}.
\item
  B Fong, D Spivak, R Tuyéras. ``Backprop as Functor: A compositional
  perspective on supervised learning.'' \emph{2019 34th Annual ACM/IEEE
  Symposium on Logic in Computer Science (LICS)} \textbf{1} (2019),
  1--13. DOI:
  \href{https://doi.org/10.1109/LICS.2019.8785665}{\texttt{10.1109/LICS.2019.8785665}}.
\item
  B Fong, D Spivak. \emph{An Invitation to Applied Category Theory:
  Seven Sketches in Compositionality}. Cambridge University Press, 2019.
  DOI:
  \href{https://doi.org/10.1017/9781108668804}{\texttt{10.1017/9781108668804}}.
  (Freely available online as
  \href{https://arxiv.org/abs/1803.05316}{arXiv:\texttt{1803.05316}}).
\item
  T Mitchell et al.~``Never-ending learning.'' \emph{Communications of
  the ACM} \textbf{61} (2018), 103--115. DOI:
  \href{https://doi.org/10.1145/3191513}{\texttt{10.1145/3191513}}.
\item
  P Selinger. ``A Survey of Graphical Languages for Monoidal
  Categories'', in \emph{New Structures for Physics}. Springer, 2010.
  Lecture Notes in Physics \textbf{813}. DOI:
  \href{https://doi.org/10.1007/978-3-642-12821-9_4}{\texttt{10.1007/978-3-642-12821-9\_4}}.
\item
  D Vagner, D Spivak, E Lerman. ``Algebras of open dynamical systems on
  the operad of wiring diagrams.'' \emph{Theory and Applications of
  Categories} \textbf{30} (2015), 1793--1822.
\end{enumerate}

\end{document}